\documentclass[10pt,twocolumn,letterpaper]{article}

\usepackage{cvpr}  
\definecolor{iccvblue}{rgb}{0.21,0.49,0.74}
\usepackage[pagebackref,breaklinks,colorlinks,allcolors=iccvblue]{hyperref}
\usepackage{graphicx}
\usepackage{amsmath}
\usepackage{amssymb}
\usepackage{colortbl}
\usepackage{graphicx}
\usepackage{xcolor}
\usepackage{colortbl}
\usepackage{multirow}
\usepackage{booktabs}
\usepackage{amssymb}
\usepackage{pifont}
\usepackage{algorithm}
\usepackage{algpseudocode}

\def\paperID{} % *** Enter the Paper ID here
\def\confName{CVPR}
\def\confYear{2026}

\title{LPT: Less-overfitting Prompt Tuning for Vision-Language Model}
\author{
Chenhao~Ding,
Xinyuan~Gao,
Songlin~Dong,
Jizhou~Han,
Qiang~Wang,
Zhengdong~Zhou,\\
Yuhang~He,
Yihong Gong, {Fellow,~IEEE}
\\
}

\begin{document}
\maketitle

\begin{abstract}
Vision-language models (VLMs) have demonstrated exceptional generalization capabilities for downstream tasks. Due to its efficiency, prompt learning has gradually become a more effective and efficient method for transferring VLMs to downstream tasks, surpassing traditional fine-tuning methods. However, during the transfer process, these models are prone to severe overfitting, leading to a significant decline in generalization ability. To address this issue, we propose a framework named LPT, specifically designed for vision-language models. Specifically, we use CLIP to filter out fine-grained foreground information that may lead to overfitting, thereby guiding the prompts with basic visual concepts. Additionally, to further mitigate overfitting, we have developed a Structural Preservation (SP) constraint at the feature level, which aligns the model’s overall feature space structure with the frozen CLIP, endowing the feature space with overall plasticity and enabling effective reshaping of the feature space during optimization. Moreover, we employ Hierarchical Logit (HL) constraint at the output layer to constrain the overall class information in the output, complementing the role of SP at the output end. Extensive experiments across various benchmarks (from base-to-novel, cross-dataset transfer, and domain generalization) demonstrate that our approach significantly improves generalization capability and effectively alleviates overfitting compared to state-of-the-art methods.
\end{abstract}

\begin{figure}[htb!]
\setlength{\abovecaptionskip}{0cm} 
\begin{center}
\centerline{\includegraphics[width=\linewidth]{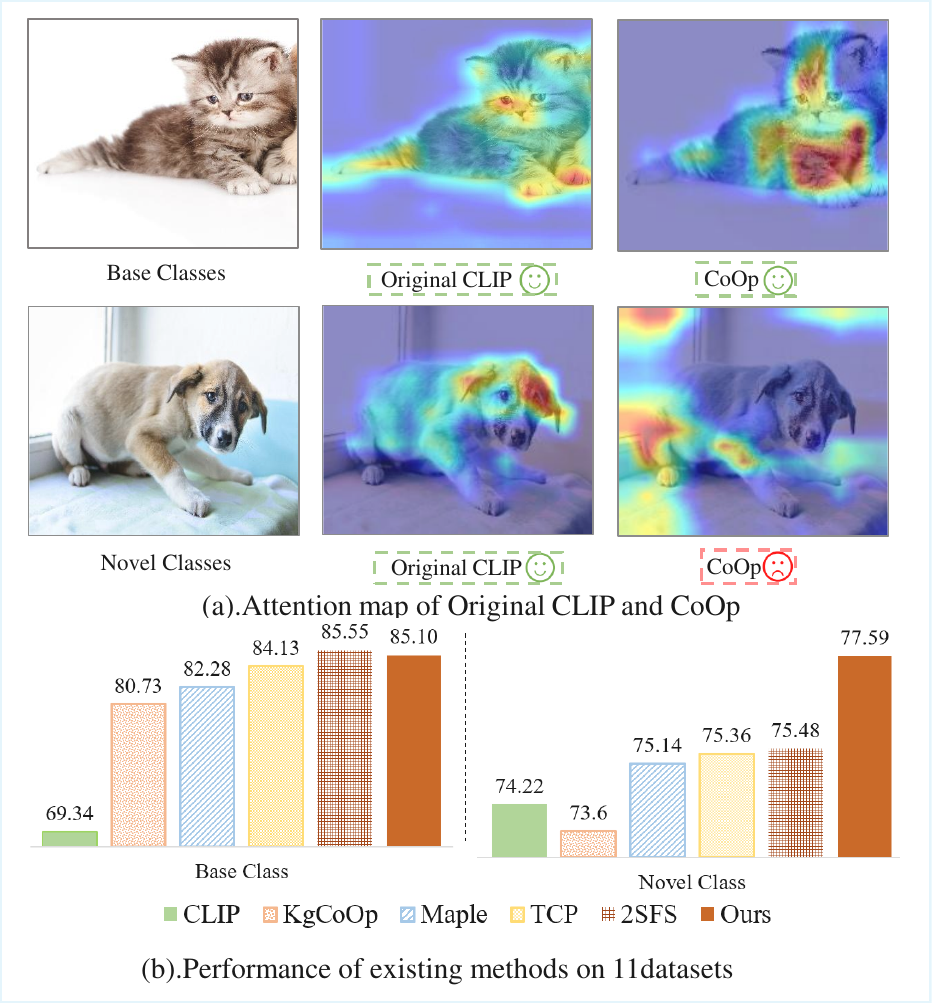}}
\caption{(a) Attention maps of CLIP and CoOp. Overfitting leads to more attention on fine details of base classes (e.g., a cat's face). The overfitted model (e.g., CoOp) struggles to transfer learned knowledge to unseen classes. (b) Performance comparison of existing methods and ours on base and novel classes. Our method achieves better improvements in novel classes while maintaining comparable performance on base classes compared to other methods.}
\label{fig:head}
\end{center}
\vspace{-5mm}
\end{figure}

\section{Introduction}
\label{intro}
{Vision}-language models (VLMs), such as CLIP~\cite{clip} and BLIP~\cite{BLIP}, have shown exceptional generalization capabilities for downstream tasks~\cite{decoupling,csvt1,csvt5}. Prompt learning has emerged as a more efficient alternative to fine-tuning VLMs, such as CoOp~\cite{coop}, introducing learnable prompt vectors to adapt models for specific downstream tasks.
However, prompts are optimized for specific tasks, prompt-based models are prone to overfitting to these tasks as training progresses, which diminishes the generalization capability of the original CLIP model for unseen tasks. Thus, maintaining generalization to unseen tasks while learning specific tasks is a crucial challenge, known as \textbf{Base to Novel}~(B2N)~\cite{coop} task.

The present mainstream approaches for the B2N task can be divided into two categories: (i) Learning class-agnostic or independent prompts to improve generalization capability~\cite{cocoop,maple}. For example, CoCoOP~\cite{cocoop} generates input-conditioned tokens for each instance rather than specific classes, thus making it less sensitive to class shift. (ii) Maintain strict consistency between the prompt model and the original CLIP to achieve the original generalization ability for novel tasks~\cite{kgcoop,tcp}. For instance, KgCoOp~\cite{kgcoop} applies knowledge distillation techniques to maintain consistency between the prompt model’s features and those of the original CLIP. Although these methods improve the generalization of the prompt model to some extent, the performance improvement for novel classes is quite limited. As shown in Fig.~\ref{fig:head} (b), the state-of-the-art method, 2FSF~\cite{2SFS}, achieves only a \textbf{0.12\%} performance improvement on unseen classes compared to the second-best method TCP. Therefore, this motivates us to explore the \textit{underlying reasons} for the performance bottleneck in B2N tasks.

We visualized the attention maps of the original CLIP model and the CoOp~\cite{coop} on base and novel class in Fig.~\ref{fig:head} (a). We can observe that the reduced zero-shot generalization ability is mainly caused by the model overfitting to the foreground details of the base class images during prompt learning. Specifically, after fine-tuning the model, CoOp tends to focus excessively on fine-grained foreground details of the image (e.g., focusing attention on specific areas of a cat) while neglecting coarse-grained structural elements (referred to as structural information), such as contours, shapes, and poses (e.g., the original CLIP model maintains a broader attention span). This structural information is crucial for reducing overfitting and overcoming performance bottlenecks in novel tasks.

Therefore, we aim to focus on the structural information rather than fine-grained foreground details during prompt learning to address the B2N problem. We propose the \textbf{L}ess-overfitting \textbf{P}rompt \textbf{T}uning~(LPT) framework, as a way for learning structural information from the images, which can effectively adapt the prompt model to unseen tasks. 

First, we propose a foreground information mask~(FIM) module, which masks fine-grained information from base class images to enable the model to focus on overall structural information rather than excessively on foreground details. Specifically, we adopt the original CLIP model to generate the attention map for each image. 
Then, we apply a suitable threshold to filter out regions with high attention, selectively removing specific foreground information from the image. Meanwhile, our study shows that this moderate refinement of fine-grained information has little impact on base class training but significantly enhances the model’s performance in novel classes.

Secondly, mainstream methods (ii) aim to restore generalization by aligning models with the original CLIP. However, these methods typically use strict distillation to maintain feature consistency, causing the model to absorb excessive fine-grained information from CLIP~(\textit{i.e.} overfit to original CILP). Given that the generalization capability of the CLIP model itself has limitations, this strict constraint approach restricts the learning capacity of prompts, thus hindering further improvements in generalization. As shown in Fig~\ref{fig:head} (b), 2SFS only improves performance on new classes by 1.26\% compared to CLIP. In contrast, we propose the structural preservation~(SP) constraint, which ensures consistency with the original CLIP only in terms of its overall structure, transferring CLIP's structural information to learn its zero-shot ability. Our research demonstrates that focusing on extracting structural information from CLIP, rather than fine-grained details, makes the feature optimization of the prompt model more flexible, thereby further enhancing the model's generalization ability to unseen tasks. This results in a significant 3.37\% improvement over CLIP on novel classes~(Fig~\ref{fig:head} (b)). Furthermore, we employ a hierarchical logit~(HL) constraint at the output layer to complement the SP constraint, ensuring the preservation of structural information while inheriting CLIP's zero-shot knowledge.

In summary, our method makes the following main contributions:
\begin{itemize}
    \item We explore the decline of generalization in B2N tasks caused by prompt model overfitting to fine-grained details and propose the LPT framework to address this problem.
    % \item We find that the decline in model generalization in B2N tasks is due to overfitting to fine-grained details.
    \item We introduce the FIM module, which masks fine-grained image details to shift the model's focus toward overall structure, thereby enhancing generalization without impacting the base class.
    \item We design SP and HL constraints, which transfer CLIP's structure information to boost generalization ability, leading to improvements on new classes compared to the original CLIP.
    \item Extensive experiments across base-to-new, cross-dataset transfer, and domain generalization demonstrate that our method effectively mitigates the overfitting problem and improves generalization capability. 
\end{itemize}

\section{Related Work}
\subsection{Vision-Language Model Pre-training}
In recent years, vision-language models have garnered widespread attention from researchers as a new and powerful tool for advancing visual recognition systems. These models, such as CLIP~\cite{clip} and BLIP~\cite{BLIP}, harness large-scale image-text pairs sourced from the web to train multimodal networks capable of understanding both visual and textual information. Through their innovative training strategies, these models establish connections between images and their associated textual descriptions, enabling them to bridge the gap between vision and language.

During pre-training, vision-language models adopt a self-supervised learning approach that aims to map images and corresponding textual descriptions into a shared embedding space. By pulling closer together the representations of matched image-text pairs and pushing apart those of mismatched pairs, these models are able to learn the nuanced semantics of both modalities in relation to one another. This method, which does not rely on manually annotated labels, enables vision-language models to acquire a robust understanding of how images and their textual counterparts are conceptually linked.

The remarkable generalization capabilities exhibited by these models have made them highly effective in a variety of downstream tasks. From object detection and image captioning to visual question answering and cross-modal retrieval, vision-language models have demonstrated impressive performance across numerous applications~\cite{nilsback2008automated,mma,2SFS,csvt3}. As a result, these models have become an essential component of modern computer vision and natural language processing pipelines, offering scalable and efficient solutions for diverse real-world problems.

\subsection{Prompt Tuning}
To enable the rapid transfer of Vision-Language Models (VLMs) to downstream tasks, several methods~\cite{coop,cocoop,maple,csvt2,csvt4} have adopted the concept of prompt tuning, which has proven effective in the Natural Language Processing (NLP) field. This approach introduces a small set of additional parameters, enabling the efficient adaptation of VLMs to new tasks without the need for extensive retraining. CoOp~\cite{coop} and CoCoOp~\cite{cocoop} achieve this by fine-tuning the frozen CLIP model through the addition of prompts to the text branch. This method allows the model to adjust to different textual inputs while keeping the visual branch fixed, thus improving task-specific performance with minimal computational overhead. 

Maple~\cite{maple} and IVLP~\cite{IVLP} extend this idea of CoOp~\cite{coop} by adding prompts to both the visual and language branches of the CLIP model, further enhancing the model’s transferability. With this dual-prompt strategy, the model can not only better adapt to different textual tasks but also effectively handle complex visual tasks, thus improving task adaptability and performance. The advantage of this approach lies in its ability to quickly adjust both visual and textual modalities to new tasks, thereby enhancing the model's performance across a wider range of applications. Meanwhile, methods such as PromptSRC~\cite{promptsrc}, KgCoOp~\cite{kgcoop}, and TCP~\cite{tcp} introduce knowledge distillation techniques to further improve the model’s generalization capability. By distilling knowledge from the frozen CLIP model into the prompts, these methods enable the model to effectively inherit useful representations learned during pre-training. In this way, knowledge distillation not only enhances the model’s transfer ability but also ensures that the model performs better in a variety of real-world applications.

However, despite the advantages of these approaches, they still face challenges in preventing the CLIP model from overfitting to base classes when transferred to downstream tasks. This overfitting problem limits the model’s ability to generalize effectively to novel classes or unseen data, which is critical for many real-world applications. Notably, some methods~\cite{argue,hpt,cpl,tree} introduce external data or large language models (LLMs) to boost the model's performance, while others~\cite{promptkd} suffer from issues such as data leakage or reliance on larger models. Due to these limitations, we have excluded comparisons with these methods in our analysis, focusing only on approaches that do not require external resources or face such challenges.

\begin{figure*}[ht]
\begin{center}
\centerline{\includegraphics[width=1\linewidth]{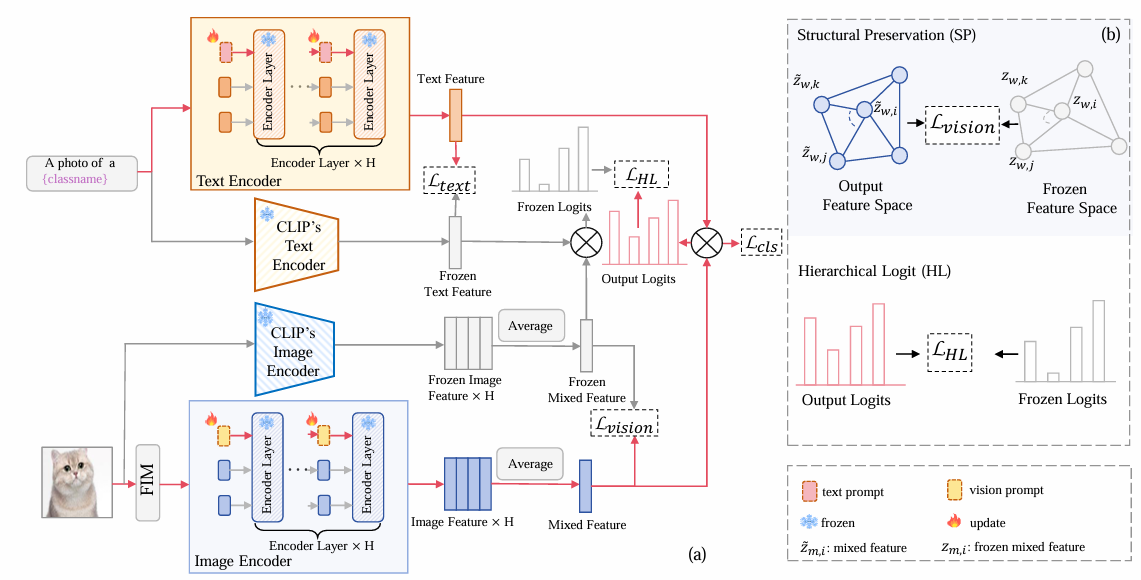}}
\caption{An overview of our method. Subfigure (a) shows the training pipeline of our method, where the text prompts and vision prompts are learnable. FIM refers to the Foreground Information Mask, which will be detailed in section \ref{sec:fim}. Subfigure (b) presents the alignment strategy of our method. By SP and HL, we preserve CLIP’s original generalization ability by maintaining its topological structure and does not constrain the prompt’s adaptability to downstream tasks.}
\label{fig:overview}
\end{center}
\end{figure*}

\subsection{Knowledge Distillation} 
% In deep learning, incorporating knowledge distillation has proven to enable student models to acquire knowledge from teacher models at a low cost~\cite{KD,mokd}. Recently, instance-level logits alignment~\cite{KD,improved}, multi-level logits alignment~\cite{mlkd}, and feature alignment \cite{cross,passalis2018learning} have become key methods for promoting knowledge transfer. In our work, however, we adapt a different constraint: by utilizing structural preservation and hierarchical information, we achieve better results for B2N task.
In deep learning, incorporating knowledge distillation has proven to be an effective strategy for enabling student models to acquire knowledge from teacher models at a low computational cost~\cite{KD,mokd}. The primary advantage of this approach lies in its ability to transfer useful knowledge, such as feature representations or decision boundaries, from a well-trained teacher model to a smaller or less complex student model. Over time, various techniques have emerged to improve the distillation process. For example, instance-level logits alignment~\cite{KD,improved} ensures that the outputs of the student model are closely aligned with those of the teacher at the individual instance level. Multi-level logits alignment~\cite{mlkd} extends this idea by aligning logits at multiple layers of the network, which allows the student model to capture more nuanced information. Additionally, feature alignment methods~\cite{cross,passalis2018learning} focus on aligning the intermediate features of the student and teacher models, further enhancing the transfer of knowledge.

While these traditional distillation techniques have been widely adopted, in our work, we take a different approach by introducing a novel constraint. Instead of merely aligning logits or features, we focus on structural preservation and the use of hierarchical information. By preserving the underlying structure of the learned representations and incorporating multi-level hierarchical knowledge, our method effectively guides the student model in transferring not just raw outputs but also high-level patterns and relationships within the data. This approach leads to better results, especially for the B2N task, where the task-specific structural information plays a crucial role in achieving superior performance. The combination of structural and hierarchical information enables a more comprehensive transfer of knowledge, ultimately improving the model's generalization capabilities and task adaptation.
\section{Experiments}
\label{sec:experiment}
%-----------------------------------------------------------
\subsection{Experimental Setup}
\noindent{\textbf{Base-to-Novel Generalization}.}
In this setup, we evaluate the generalization of LPT. Following CoOp~\cite{coop}, we divide the dataset evenly into two parts: base classes and novel classes. We conduct 16-shot learning on the base classes and perform zero-shot evaluations on the novel classes. This setup aims to evaluate the model's generalization capability when transferring to downstream tasks.

\noindent{\textbf{Cross-Dataset Transfer}.}
Following the approach of CoOp~\cite{coop}, in \emph{cross-dataset transfer}, we conduct 16-shot few-shot training on ImageNet1k~\cite{imagenet}, and then perform zero-shot testing on multiple heterogeneous datasets. Through cross-dataset transfer testing, we are able to evaluate the model's generalization and robustness across datasets.

\begin{table*}[!h] 
\caption{Compare the performance on novel classes of our method with existing methods on base-to-novel generalization. }
    \centering
    \resizebox{\linewidth}{!}{
    \begin{tabular}{l c ccccccccccc}
    \toprule
    \fontsize{13pt}{9pt}\selectfont
    &  & \multicolumn{11}{c}{\textbf{Novel}} \\ \cmidrule(lr){2-13}
    & \rotatebox{45}{ImageNet} &\rotatebox{45}{Caltech101} & \rotatebox{45}{OxfordPets} & \rotatebox{45}{StanfordCars} & \rotatebox{45}{Flowers102} & \rotatebox{45}{Food101} & \rotatebox{45}{Aircraft} &\rotatebox{45}{SUN397} &\rotatebox{45} {DTD} & \rotatebox{45}{EuroSAT} &\rotatebox{45} {UCF101} & \rotatebox{45}{\emph{Average} }\\
    \midrule 
    CLIP~\cite{clip} & 68.14 & 94.00 & 97.26 & 74.89 & \underline{77.80} & 91.22 & 36.29 & 75.35 & 59.90 & 64.05 & 77.50 & 74.22\\
    
    CoOp~\cite{coop} & 67.88 & 89.81 & 95.29 & 60.40 & 59.67 & 82.26 & 22.30 & 65.89 & 41.18 & 54.74 & 56.05 & 63.22 \\
    
    CoCoOp~\cite{cocoop} & 70.43 & 93.81 & 97.69 & 73.59 & 71.75 & 91.29 & 23.71 & 76.86 & 56.00 & 60.04 & 73.45 & 71.69 \\
    
    LASP~\cite{lasp} & 70.95 & 94.24 & \underline{97.93} & 71.60 & 74.00 & \underline{91.70} & 30.57 & 78.60 & 58.60 & 77.78 & 78.03 & 74.90 \\
    
    MaPLe~\cite{maple} & 70.54 & 94.36 & 97.76 & 74.00 & 72.46 & \textbf{92.05} & 35.61 & 78.70 & 59.18 & 73.23 & 78.66 & 75.14 \\
    
    PromptSRC~\cite{promptsrc} & 70.73 & 94.03 & 97.30 & \underline{74.97} & 76.50 & 91.53 & \underline{37.87} & 78.47 & 62.97 & 73.90 & 78.80 & 76.10 \\
    
    Dept~\cite{dept} & 70.13 & \underline{94.60} & 97.23 & \textbf{75.47} & 76.37 & 91.60 & 34.83 & 77.80 & 59.13 & 71.07 & 77.23 & 75.04  \\
    
    TCP~\cite{tcp} & 69.87 & \textbf{94.67} & 97.20 & 74.13 & 75.57 & 91.37 & 34.43 & 78.20 & 58.07 & 74.73 & \textbf{80.77} & 75.36  \\
    
    MMA~\cite{mma} & \underline{71.00} & 94.00 & \textbf{98.07} & 73.10 & 75.93 & 91.30 & 36.33 & 78.57 & \textbf{65.63} & \textbf{82.34} & 80.03 & \underline{76.80} \\
    
    2SFS~\cite{2SFS} & 70.99 & 94.43 & 97.82 & 74.80 & 76.17 & 91.34 & 35.51 & \underline{78.91} & 65.01 & 67.09 & 78.19 & 75.48 \\
    \midrule
    \textbf{LPT (Ours)} & \textbf{71.70} & 94.30 & 97.87 & \textbf{75.47} & \textbf{77.90} & 91.67 & \textbf{38.80} & \textbf{79.50} & \underline{65.40} & \underline{80.37} & \underline{80.50} & \textbf{77.59} \\
    \bottomrule
    \end{tabular}}
        
    \label{tab:novel}
\end{table*}
\begin{table*}[!h]
 \caption{Compare the performance on base classes of our method with existing methods on base-to-novel generalization. }
    \centering
    \resizebox{\linewidth}{!}{
    \begin{tabular}{l c ccccccccccc}
    \toprule
    \fontsize{13pt}{9pt}\selectfont
    &  & \multicolumn{11}{c}{\textbf{Base}} \\ \cmidrule(lr){2-13}
    & \rotatebox{45}{ImageNet} &\rotatebox{45}{Caltech101} & \rotatebox{45}{OxfordPets} & \rotatebox{45}{StanfordCars} & \rotatebox{45}{Flowers102} & \rotatebox{45}{Food101} & \rotatebox{45}{Aircraft} &\rotatebox{45}{SUN397} &\rotatebox{45} {DTD} & \rotatebox{45}{EuroSAT} &\rotatebox{45} {UCF101} & \rotatebox{45}{\emph{Average} }\\
    \midrule 
    CLIP~\cite{clip} & 72.43 & 96.84 & 91.17 & 63.37 & 72.08 & 90.10 & 27.19 & 69.36 & 53.24 & 56.48 & 70.53 & 69.34\\
    CoOp~\cite{coop} & 76.47 & 98.00 & 93.67 & 78.12 & 97.60 & 88.33 & 40.44 & 80.60 & 79.44 & 92.19 & 84.69 & 82.69 \\
     CoCoOp~\cite{cocoop} & 75.98 & 97.96 & 95.20 & 70.49 & 94.87 & 90.70 & 33.41 & 79.74 & 77.01 & 87.49 & 82.33 & 80.47\\
    LASP~\cite{lasp} & 76.20 & 98.10 & \textbf{95.90} & 75.17 & 97.00 & \textbf{91.20} & 34.53 & 80.70 & 81.40 & 94.60 & 84.77 & 82.70 \\
    MaPLe~\cite{maple} & 76.66 & 97.74 & 95.43 & 72.94 & 95.92 & 90.71 & 37.44 & 80.82 & 80.36 & 94.07 & 83.00 & 82.28 \\
    PromptSRC~\cite{promptsrc} & 77.60 & 98.10 & 95.33 & 78.27 & \underline{98.07} & 90.67 & 42.73 & \underline{82.67} & \underline{83.37} & 92.90 & 87.10 & 84.26 \\	
    Dept~\cite{dept} & 77.03 & 98.30 & 94.33 & 79.13 & 98.00 & 90.50 & 43.20 & 82.33 & 82.20 & 89.03 & 85.80 & 83.62 \\
    TCP~\cite{tcp} & 77.27 & 98.23 & 94.67 & \underline{80.80} & 97.73 & 90.57 & 41.97& 82.63 & 82.77 & 91.63 & 87.13 & 84.13  \\
    MMA~\cite{mma} & 77.31 & 98.40 & 95.40 & 78.50 & 97.77 & 90.13 & 40.57 & 82.27 & 83.20 & 85.46 & 86.23 & 83.20 \\
    2SFS~\cite{2SFS} & \underline{77.71} & \underline{98.71} & 95.32 & \textbf{82.50} & \textbf{98.29} & 89.11 & \textbf{47.48} & 82.59 & \textbf{84.60} & \textbf{96.91} & \textbf{87.85} & \textbf{85.55} \\
    \midrule
    \textbf{LPT (Ours)} & \textbf{78.53} & \textbf{98.73} & \underline{95.60} & 80.10 & 97.47 & \underline{90.73} & \underline{47.20} & \textbf{82.83} & 81.40 & \underline{95.80} & \underline{87.73} & \underline{85.10} \\										
    \bottomrule
    \end{tabular}}
        
    \label{tab:base}
\end{table*}

\noindent{\textbf{Domain Generalization}.}
We evaluated the robustness of our method on out-of-distribution datasets. Similar to the \emph{cross-dataset} evaluation, we first perform few-shot learning on ImageNet1K~\cite{imagenet} and then test our ImageNet-trained model on four other ImageNet datasets that involve different types of domain shifts, including ImageNet-A~\cite{imageneta}, ImageNet-R~\cite{imagenetr}, ImageNet-V2~\cite{imagenetv2}, and ImageNet-S~\cite{imagenetsk}.
% \noindent{\textbf{Domain Generalization}.}This setup aims to test the model's performance on OOD (out-of-distribution) datasets. In Domain Generalization, we conduct few-shot training on ImageNet1k~\cite{imagenet} and then perform zero-shot testing on various ImageNet variant datasets.
\begin{table*}[!h]
    \caption{Performance comparison on cross-dataset transfer evaluation. Our method achieves the highest accuracy on 6 datasets and the second-best on 3, yielding the best overall average. The best and second-best results are indicated in \textbf{bold} and \underline{underlined}, respectively. The model capacity remains consistent across all methods.}
    \centering
    \small
    \resizebox{\linewidth}{!}{
    \begin{tabular}{l c ccccccccccc}
    \toprule
    \fontsize{13pt}{9pt}\selectfont
    & \textbf{Source} & \multicolumn{11}{c}{\textbf{Target}} \\ \cmidrule(lr){2-2} \cmidrule(lr){3-13}
    & \rotatebox{45}{ImageNet} & \rotatebox{45}{Caltech101} & \rotatebox{45}{OxfordPets} & \rotatebox{45}{StanfordCars} & \rotatebox{45}{Flowers102} & \rotatebox{45}{Food101} & \rotatebox{45}{Aircraft} &\rotatebox{45}{SUN397} & \rotatebox{45}{DTD} & \rotatebox{45}{EuroSAT} & \rotatebox{45}{UCF101} & \rotatebox{45}{\emph{Average}} \\
    \midrule
    CLIP~\cite{clip} & 66.70 & 93.70 & 89.10 & 65.70 & 70.70 & 85.90 & 24.90 & 62.60 & 44.30 & 48.30 & 67.60 & 65.24\\
    CoOp~\cite{coop} & \underline{71.51} & 93.70 & 89.14 & 64.51 & 68.71 & 85.30 & 18.47 & 64.15 & 41.92 & 46.39 & 66.55 & 63.88 \\
    CoCoOp~\cite{cocoop} & 71.02 & \underline{94.43} & 90.14 & 65.32 & 71.88 & 86.06 & 22.94 & 67.36 & 45.73 & 45.37 & 68.21 & 65.74 \\
    %  MaPLe~\cite{maple} & 70.72 & 93.53 & {90.49} & {65.57} & \textbf{72.23} & {86.20} & {24.74} & {67.01} & {46.49} & {48.06} & {68.69} & \underline{66.30} \\
    PromptSRC~\cite{promptsrc} & 71.27 & 93.60 & 90.25 & 65.70 & 70.25 & 86.15 & 23.90 & 67.10 & \textbf{46.87} & 45.50 & \underline{68.75} & 65.81 \\
    TCP~\cite{tcp} & 71.40 & 93.97 & \textbf{91.25} & 64.69 & 71.21 & \textbf{86.69} & 23.45 & 67.15 & 44.35 & \underline{51.45} & 68.73 & 66.29 \\
    MMA~\cite{mma} & 71.00 & 93.80 & 90.30 & \underline{66.13} & \underline{72.07} & 86.12 & \underline{25.33} & \textbf{68.17} & \underline{46.57} & 49.24 & 68.32 & \underline{66.61} \\
    \midrule
    \textbf{LPT (Ours)} & \textbf{71.77} & \textbf{95.00} & \underline{90.87} & \textbf{66.50} & \textbf{73.40} & \underline{86.53} & \textbf{25.67} & \underline{68.00} & 43.95 & \textbf{54.13} & \textbf{69.27} & \textbf{67.33} \\
    \bottomrule
    \end{tabular}}
  
    \label{tab:crossdata}
\end{table*}

\noindent{\textbf{Datesets}.}
For \emph{base-to-movel generalization} and \emph{cross dataset transfer} setting, we use ImageNet~\cite{imagenet}, FGVCAircraft~\cite{aircraft}, EuroSAT~\cite{eurosat}, UCF101~\cite{ucf101}, DTD~\cite{dtd}, Caltech101~\cite{caltech101}, Oxford-Pets~\cite{pets}, Stanford-Cars~\cite{cars}, Oxford-Flowers~\cite{flowers}, Food101~\cite{food101}, and SUN397~\cite{sun397}. For \emph{domain generalization}, we use ImageNet1k~\cite{imagenet} as the source domain and ImageNet-A~\cite{imageneta}, ImageNet-R~\cite{imagenetr}, ImageNet-V2~\cite{imagenetv2}, and ImageNet-Sketch~\cite{imagenetsk} as the target domains. For \emph{base-to-novel generalization}, based on the performance of CoOp, we classify DTD~\cite{dtd}, FGVCAircraft~\cite{aircraft}, EuroSAT~\cite{eurosat}, and UCF101~\cite{ucf101} as challenging datasets. These four datasets have fine-grained characteristics, making them more prone to overfitting compared to other datasets. For \emph{base-to-novel generalization}, based on the performance of zero-shot CLIP, we classify DTD~\cite{dtd}, FGVCAircraft~\cite{aircraft}, EuroSAT~\cite{eurosat}, and UCF101~\cite{ucf101} as challenging datasets. These four datasets have fine-grained characteristics, making them more prone to overfitting compared to other datasets. 

\noindent{\textbf{Implementation Details}.}
Our implementation is based on the ViT-B/16 variant of the CLIP model. We trained for 20 epochs across all three benchmarks. For Base-to-Novel Generalization, prompts were added in the first 7 transformer layers, while for the other two benchmarks, this number was 3. We employed the Adam optimizer with a learning rate set to 2.5e-3. Performance metrics included base class accuracy, novel class accuracy, and harmonic mean accuracy, averaged over 3 experimental runs to determine final accuracy. Our experiments were conducted on an RTX 3090 24G. All prompts are initialized randomly from a normal distribution, except for the text prompt in the first layer, which is initialized with the word embeddings of “X X X X.” Independent Vision-Language Prompting (IVLP)~\cite{IVLP} was used as the baseline method. In terms of hyperparameter settings, we set $\gamma$ to 3, $\lambda$ to 1, and for the mask threshold in FIM, we set it to 30 to remove high-attention regions in the image.
% ImageNet1k~\cite{imagenet}, FGVCAircraft~\cite{aircraft}, EuroSAT~\cite{eurosat}, UCF101~\cite{ucf101}, DTD~\cite{dtd}, Caltech101~\cite{caltech101}, Oxford-Pets~\cite{pets}, Stanford-Cars~\cite{cars}, Oxford-Flowers~\cite{flowers}, Food101~\cite{food101}, and SUN397~\cite{sun397}. For \emph{domain generalization}, we used ImageNet1k~\cite{imagenet} as the source domain and ImageNet-A~\cite{imageneta}, ImageNet-R~\cite{imagenetr}, ImageNet-V2~\cite{imagenetv2}, and ImageNet-Sketch~\cite{imagenetsk} as the target domains.
%-----------------------------------------------------------

%-----------------------------------------------------------
\subsection{Comparison Methods}
We compare our method with several representative baselines: CLIP~\cite{clip}, CoOp~\cite{coop}, CoCoOp~\cite{cocoop}, LASP~\cite{lasp}, MaPLe~\cite{maple}, PromptSRC~\cite{promptsrc}, Dept~\cite{dept}, TCP~\cite{tcp}, MMA~\cite{mma}, and 2SFS~\cite{2SFS}. Among these, CoOp, CoCoOp, LASP, Dept, and TCP are text-prompting methods that introduce learnable prompts exclusively within the text encoder. In contrast, MaPLe, PromptSRC, and our method employ multi-modal prompting by incorporating prompts into both the visual and text encoders. Additionally, MMA and 2SFS are included as representative adapter-based approaches.

\subsection{Base-to-Novel Generalization}
We compare our method with several representative baselines: zero-shot CLIP~\cite{clip}, CoOp~\cite{coop}, CoCoOp~\cite{cocoop}, LASP~\cite{lasp}, MaPLe~\cite{maple}, PromptSRC~\cite{promptsrc}, Dept~\cite{dept}, TCP~\cite{tcp}, MMA~\cite{mma}, and 2SFS~\cite{2SFS}. Among these, CoOp, CoCoOp, LASP, Dept, and TCP are text-prompting methods that introduce learnable prompts exclusively within the text encoder. In contrast, MaPLe, PromptSRC, and our method employ multi-modal prompting by incorporating prompts into both the visual and text encoders. Additionally, MMA and 2SFS are included as representative adapter-based approaches. Table \ref{tab:base2novel} presents the results across 11 datasets. Additionally, based on the performance of zero-shot CLIP and the fine-grained nature of the datasets, we classify FGVCAircraft, EuroSAT, DTD, and UCF101 as challenging datasets. Their overall accuracy is shown in Table \ref{tab:novel}. Overall, all existing methods outperform CLIP on base classes, but their performance on novel classes shows only limited improvement compared to CLIP and remains stagnant. The model size used by all comparison methods is consistent.

The stagnation in novel class performance suggests that existing methods overfit to the fine-grained details of base classes, neglecting overall information, which weakens CLIP's ability to capture the fundamental visual features required for effective transfer. In contrast, our method better captures underlying visual structure through foreground information filtering and structure preservation, allowing the model to focus more on overall information rather than fine-grained details. This provides greater flexibility in the feature space, reduces overfitting, and enhances the model's generalization capability. 

\begin{table} 
  \caption{\textnormal{Domain generalization. }Prompt learning methods are trained on ImageNet and evaluated on datasets with domain shifts. The best results are in \textbf{bold} and the second-best results are \underline{underlined}. The model size used by all comparison methods is consistent.} 
    \small \centering
 \setlength{\tabcolsep}{1.2mm}
 \selectfont
    {
    \begin{tabular}{l cccccc}
    \toprule
    & \textbf{Source} & \multicolumn{5}{c}{\textbf{Target}} \\ \cmidrule(lr){2-2} \cmidrule(lr){3-7}
     & ImageNet & -V2 & -S & -A & -R  & Avg.\\
    \midrule
    CLIP~\cite{clip} &  66.73 & 60.83 & 46.15 & 47.77 & 73.96 & 57.18 \\
    CoOp~\cite{coop} &  71.51 & 64.20 & 47.99  & 49.71  & 75.21  & 59.28 \\
    CoCoOp~\cite{cocoop} & 71.02 & 64.07 & 48.75 & 50.63 & 76.18 & 59.91  \\
    MaPLe~\cite{maple} & 70.72  & 64.07 & 49.15  & 50.90 & 76.98 & 60.27  \\
    DAPT~\cite{DAPT} & \underline{71.67} & \underline{64.50} & \textbf{49.53} & 51.10 & 76.33 & 60.37 \\
    TCP~\cite{tcp} & 71.20 & \textbf{64.60} & \underline{49.50} & \underline{51.20} & 76.73 & \underline{60.51} \\
    MMA~\cite{mma} & 71.00 & 64.33 & 49.13 & 51.12 & \underline{77.32} & 60.48 \\
    \midrule
    \textbf{LPT~(Ours)} & \textbf{71.77} & 64.40 & 49.47 & \textbf{51.67} & \textbf{77.83} & \textbf{60.84}\\
    \bottomrule
    \end{tabular}}
    \label{tab:domain}
\end{table}
\begin{table*}[!t]
    \caption{\small Detailed performance comparison on 4 challenging datasets showing the effect of individual components of our LPT. Absolute gains of LPT (baseline + $\mathcal{L}_{\mathrm{SP}}$ + $\mathcal{L}_{\mathrm{HL}}$ + FIM) over the baseline are shown in \textcolor{MidnightBlue}{blue}. }
    \small \centering
 \setlength{\tabcolsep}{12pt}
    \scalebox{0.9}[0.9]{
\begin{tabular}{lc|cccc|c}
\toprule
\textbf{Dataset} & 
 & 
\textbf{baseline} &
\textbf + $\mathcal{L}_\mathrm{SP}$ & 
\textbf + $\mathcal{L}_\mathrm{SP}$ + $\mathcal{L}_\mathrm{HL}$ & 
\textbf + $\mathcal{L}_\mathrm{SP}$ + $\mathcal{L}_\mathrm{HL}${ + FIM} &
\textbf{$\Delta$} \\ \midrule
% \scriptsize{$\pm$0.76}
\multirow{3}{*} {{Average}}  & Base Acc.& 73.77 & 75.82 & 78.65 & 78.03 &   \textcolor{MidnightBlue}{{+4.26}}\\
                               & Novel Acc.     & 54.63  & 62.45 & 64.52 & 66.27  &   \textcolor{MidnightBlue}{{+11.64}}\\
                               & H.M            & 62.77  & 68.49  & 70.89 & 71.62  &   \textcolor{MidnightBlue}{{+8.85}}\\
\midrule
\multirow{3}{*}{FGVCAircraft}  & Base Acc.       & 39.60 & 43.50 & 48.00 & 47.20  &  \textcolor{MidnightBlue}{{+7.60}}\\
                               & Novel Acc.      & 25.23 & 33.50  & 36.10 & 38.80   &  \textcolor{MidnightBlue}{{+13.57}}\\
                               & H.M             & 30.82 & 37.85  & 41.21  & 42.59  &  \textcolor{MidnightBlue}{{+11.77}}\\
\midrule
\multirow{3}{*}{DTD}           & Base Acc.       & 80.40 & 81.00  & 82.30 & 81.40   &  \textcolor{MidnightBlue}{{+1.00}}\\
                               & Novel Acc.      & 56.20 & 60.80  & 63.50 & 65.40  &  \textcolor{MidnightBlue}{{+9.20}}\\
                               & H.M            & 66.16 & 69.46   & 71.69 & 72.53  & \textcolor{MidnightBlue}{{+6.37}}\\
\midrule
\multirow{3}{*}{EuroSAT}       & Base Acc.       & 90.13 & 92.50  & 95.60 & 95.80 &  \textcolor{MidnightBlue}{{+5.67}}\\
                               & Novel Acc.      & 62.90 & 77.00  & 78.80 & 80.37  & \textcolor{MidnightBlue}{{+17.47}}\\
                               & H.M            & 74.09 & 84.04   & 86.39 & 87.41   &  \textcolor{MidnightBlue}{{+13.32}}\\
\midrule
\multirow{3}{*}{UCF101}        & Base Acc.      & 84.93  & 86.28  & 88.70 & 87.73   & \textcolor{MidnightBlue}{{+2.80}}\\
                               & Novel Acc.      & 74.17  & 78.50 & 79.68 & 80.50  &  \textcolor{MidnightBlue}{{+6.33}}\\
                               & H.M            & 79.62  & 82.21  & 83.95 & 83.96  & \textcolor{MidnightBlue}{{+4.34}}\\
\bottomrule
\end{tabular}%
}

    \label{tab:abs_base_to_new}
\end{table*}
\begin{table}
\captionof{table}{Effectiveness of our proposed techniques. Results are averaged over 4 challenging datasets. HM stands for harmonic mean.}
\small
\centering
\setlength{\tabcolsep}{1.8mm}
\begin{tabular}{lccccccl}
\toprule
\multirow{2}{*}{Methods} & \multirow{2}{*}{SP} & \multirow{2}{*}{HL} & \multirow{2}{*}{FIM} & \multicolumn{3}{c}{Acc(\%)} \\
\cline{5-7} 
& & & & Base & Novel & HM  \\
\midrule
baseline & & & & 73.77 & 54.63 & 62.77 \\
\quad+SP & \checkmark & & & 75.82 & 62.45 & 68.49 \\
\quad+HL & &\checkmark & & 75.95 & 60.85 & 67.57 \\
\quad+FIM & & &\checkmark & 72.84 & 61.41 & 66.64 \\
\quad+SP+HL & \checkmark & \checkmark &  & \textbf{78.65} & 64.52 & 70.89  \\
\quad+SP+HL+FIM & \checkmark & \checkmark & \checkmark & 78.03 & \textbf{66.27} & \textbf{71.62}\\
\bottomrule
\end{tabular}
\label{tb:abalation}
\end{table}

Table \ref{tab:base2novel} and Table \ref{tab:novel} demonstrate that our method outperforms 2SFS on the Base-to-Novel task, with improvements of \textbf{4.82\%} and \textbf{2.57\%} on novel classes and harmonic mean across the four challenging datasets. Across all eleven datasets, we achieve the second-best performance of 85.10\% on base classes, just 0.45\% behind 2SFS. Most notably, we achieve the best results of 77.59\% and 81.02\% on novel classes and harmonic mean, surpassing 2SFS by \textbf{2.11\%} and \textbf{0.82\%}, respectively. The experimental results further support our view that focusing on overall information, rather than fine-grained details, is more beneficial for improving the model's generalization capability.

\subsection{Cross-Dataset Transfer}
Table \ref{tab:crossdata} presents the results of our cross-dataset transfer experiments. On the source dataset ImageNet, our method achieves the highest accuracy of \textbf{71.77\%}, outperforming the competitive adapter-based method MMA~\cite{mma} (71.00\%). More importantly, when transferring to unseen target domains, our approach demonstrates superior generalization capability. Specifically, our method achieves the best performance on 6 out of 10 target datasets and the second-best on 3. Compared directly to MMA, LPT achieves higher accuracy on 8 out of 10 target datasets. Overall, LPT yields the highest average target accuracy of \textbf{67.33\%}, leading MMA by \textbf{0.72\%}. These results indicate that our method is highly advantageous for cross-dataset zero-shot transfer. Furthermore, it validates the effectiveness of our proposed designs (i.e., FIM, along with SP and HL constraints) in mitigating model overfitting to the source distribution. By focusing on overall structural information rather than dataset-specific fine-grained details, our model achieves more robust generalization across diverse target domains.

\subsection{Domain Generalization}
Table \ref{tab:domain} shows the performance of our method compared to previous methods on out-of-distribution (OOD) datasets. Following the standard protocol, we conduct 16-shot training on the source ImageNet dataset and then evaluate directly on four out-of-distribution target datasets: ImageNet-V2, ImageNet-Sketch (-S), ImageNet-A, and ImageNet-R. On the source dataset, our method achieves the highest accuracy of \textbf{71.77\%}. Furthermore, compared to the strong baseline MMA~\cite{mma}, our method yields consistent improvements across all four target datasets, resulting in an increase of \textbf{0.36\%} in average target accuracy (60.84\% vs. 60.48\%). Specifically, LPT achieves the best performance on ImageNet-A (51.67\%) and ImageNet-R (77.83\%), leading to the highest overall average among all methods. Even in the face of out-of-distribution scenarios, our method effectively transfers to datasets with significant domain shifts. The performance on the Domain Generalization benchmark further demonstrates that our method, by focusing on overall information rather than overfitting to specific domains, achieves better generalization and cross-distribution robustness.

\subsection{Ablation Studies and Analyze}
%----------------------------------------------------------

\noindent\textbf{Effectiveness of different components.} We conducted ablation experiments to evaluate the individual and progressive contributions of the Foreground Information Mask (FIM), SP, and HL in preventing model overfitting and alleviating the B2N problem. Tables \ref{tb:abalation} and \ref{tab:abs_base_to_new} show the effectiveness of each component.

We observed that the baseline model performs well on base classes (73.77\%) but poorly on novel classes (54.63\%), indicating that the baseline model has overfitted to the base classes, which compromises the original generalization ability of CLIP. After applying SP, performance on novel classes improved significantly by \textbf{7.82\%} (reaching 62.45\%), demonstrating that by learning overall structural information rather than focusing on individual image details, the model achieves better generalization. By further introducing HL, the model not only maintained the improvement in novel class performance (increasing by another \textbf{2.07\%} to 64.52\%) but also significantly enhanced base class performance by \textbf{2.83\%} (reaching the highest base accuracy of 78.65\%). This indicates that by applying class-level perceptual constraints, the model can learn more category-general knowledge, thus promoting task transfer and generalization. Finally, by incorporating FIM, we removed some foreground detail information, forcing the model to focus on underlying visual features. This further enhanced generalization, improving novel class performance by \textbf{1.75\%} (reaching 66.27\%) with minimal impact on base class learning (a slight drop of 0.62\%). These experiments validate our perspective on the overfitting issue in B2N tasks: by learning \textbf{overall structural information} rather than foreground details, the model can \textbf{capture more general knowledge}, leading to improved transferability and generalization.

\begin{table}
    \captionof{table}{Effectiveness of structural preservation constraint on 4 challenging datasets. Our structural preservation constraint provides better performance compared to traditional distance metrics within the LPT framework.}
    \small\centering
    \setlength{\tabcolsep}{8pt}
    \begin{tabular}{l ccc}
    \toprule
    Method  & Base Acc. & Novel Acc. & HM\\
    \midrule
    baseline & 73.77 & 54.63 & 62.77 \\
    
    1. LPT w $\mathcal{L}_1$ & 75.32 & 62.15 & 68.10 \\      
    2. LPT w $\mathcal{L}_2$ & 75.81 & 61.42 & 67.86 \\      
    3. LPT w SP (Ours) & \textbf{78.03} & \textbf{66.27} & \textbf{71.62} \\
    \bottomrule
    \end{tabular}
    \label{tab:ablations_sp_comparison}
\end{table}
\begin{table}
    \captionof{table}{Effectiveness of hierarchical logit constraint on 4 challenging datasets. Our hierarchical logit constraint provides better performance compared to traditional KL divergence within the LPT framework.}
   \small\centering
  \setlength{\tabcolsep}{8pt}
    {
    \begin{tabular}{l cccc}
    \toprule
    Method  & Base Acc. & Novel Acc. & HM\\
    \midrule        
    baseline & 73.77 & 54.63 & 62.77 \\
            
    1: LPT w $\mathcal{D}_{KL}$ & 77.51 & 63.59 & 70.11 \\
    2: LPT w HL (Ours) & \textbf{78.03} & \textbf{66.27} & \textbf{71.62} \\
    \bottomrule
    \end{tabular}
    }
    \label{tab:ablations_on_KD}
\end{table}

\begin{figure*}[!t]
    \centering
    \centerline{\includegraphics[width=1\linewidth]{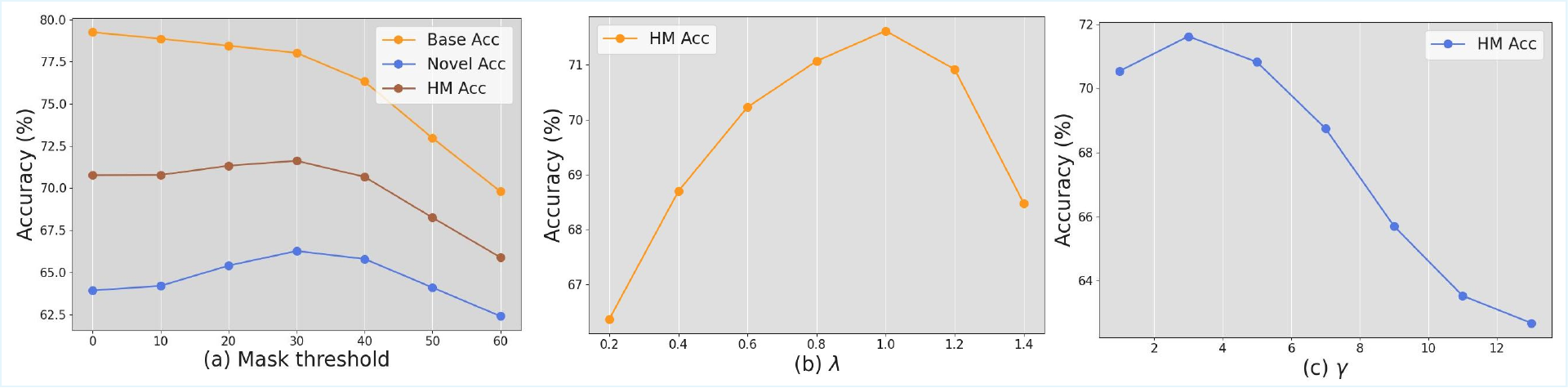}}
    \caption{The effect of mask thresholds and sensitivity analysis of $\gamma$ and $\lambda$. We report the average results across four challenging datasets.}
    \label{fig:ablations_onthers}
\end{figure*}
\begin{figure}
\begin{center}
\centerline{\includegraphics[width=\linewidth]{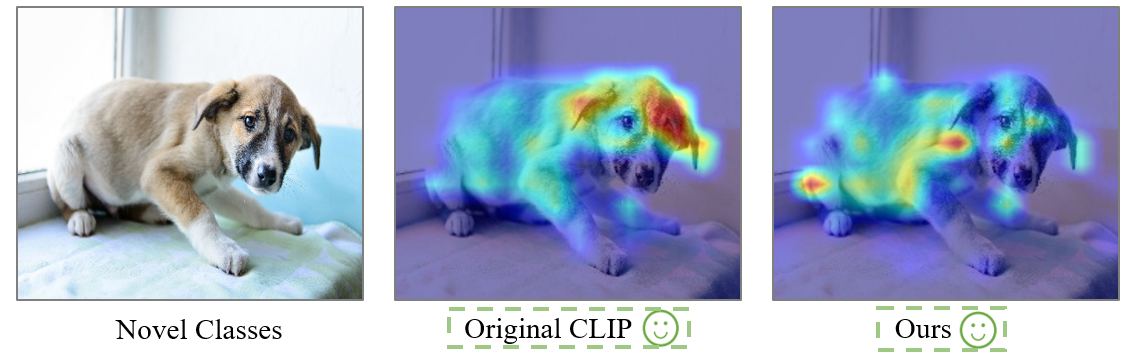}}
\caption{Attention maps of CLIP and ours on novel classes. Compared to CoOp, our method better preserves the original CLIP's overall focus in zero-shot scenarios and performs fine-tuning on the attention.}
\label{fig:attours}
\vspace{-3mm}
\end{center}
\end{figure}

\noindent\textbf{Analysis of feature space constraint methods.} Table \ref{tab:ablations_sp_comparison} presents a comparison of different feature space constraint strategies within the LPT framework. The results clearly demonstrate that our proposed Structural Preservation (SP) constraint significantly outperforms traditional distance-based metrics ($\mathcal{L}_1$ and $\mathcal{L}_2$) in terms of generalization. While the $\mathcal{L}_1$ and $\mathcal{L}_2$ constraints achieve novel class accuracies of 62.15\% and 61.42\% respectively, SP surpasses them by a substantial margin, reaching 66.27\%. This indicates that instead of rigidly restricting feature representations like traditional distance metrics, our method endows the feature space with greater plasticity. This flexibility effectively mitigates overfitting to the source distribution and facilitates superior transferability to downstream tasks, ultimately allowing SP to achieve the highest harmonic mean accuracy of 71.62\%.

\noindent\textbf{Effectiveness of foreground information mask.}
The effectiveness of foreground information masking is illustrated in Figure \ref{fig:ablations_onthers}(a), which demonstrates the impact of reducing foreground information masking on the model. Our findings indicate that the removal of a small portion of foreground information has a minimal effect on the model's ability to learn basic classes, with almost negligible impact. Furthermore, through the use of foreground information masking, the model acquires a \textbf{greater understanding of underlying visual information}, which enhances its generalization capabilities to unseen novel classes.

\noindent\textbf{Analysis of logits constraint.} Table \ref{tab:ablations_on_KD} compares the performance of different logit constraint methods within the LPT framework. By applying the proposed Hierarchical Logit (HL) constraint, our method consistently outperforms the traditional KL divergence ($\mathcal{D}_{KL}$) across all metrics. Specifically, the HL constraint yields an improvement of \textbf{0.52\%} on base classes and a significant \textbf{2.68\%} on novel classes, culminating in a \textbf{1.51\%} higher harmonic mean. This substantial gain on unseen classes indicates that the hierarchical logit constraint effectively guides the model to learn more holistic, category-general information. Consequently, it alleviates the tendency to overfit to the source distribution, \textbf{promoting better transferability than standard knowledge distillation techniques}.

\begin{figure}[t]
\begin{center}
\centerline{\includegraphics[width=0.9\linewidth]{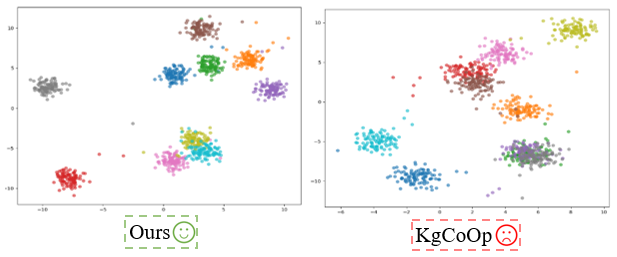}}
\caption{T-SNE visualization comparison with KgCoOp's novel class feature space.}
\label{fig:ft_space}
\end{center}
\vspace{-3mm}
\end{figure}

\noindent\textbf{Comparison with strict CLIP alignment methods.} 
As shown in Figure \ref{fig:ft_space}, due to over-aligning with CLIP, KgCoOp struggles to acquire downstream task knowledge, leading to a crowded new class feature space. In contrast, our method balances learning CLIP’s structural information while effectively integrating downstream task knowledge, allowing better differentiation of unseen classes in downstream tasks.

\begin{figure}
\begin{center}
\centerline{\includegraphics[width=1\linewidth]{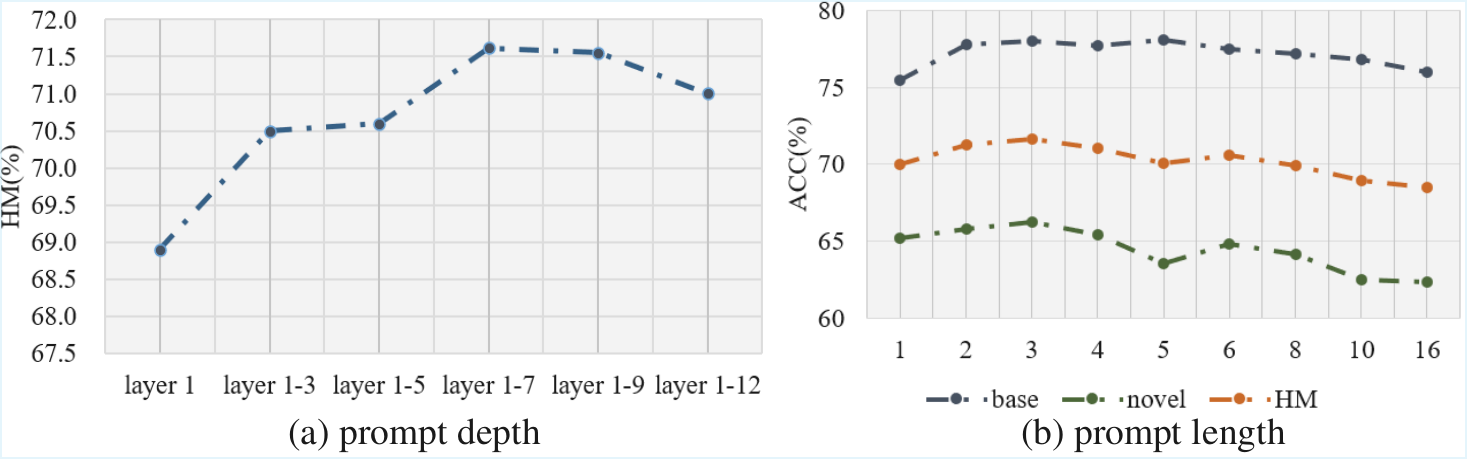}}
\caption{Ablation study on prompt depth (left) and prompt length (right) in LPT. We report the average results across four challenging datasets.}
\label{fig:dpth}
\end{center}
\vspace{-5mm}
\end{figure}

\noindent\textbf{Visualization of attention map.}
To further validate the effectiveness of our method in learning new classes within the base-to-novel task, we extracted the attention maps from the final layer of the ViT in the CLIP image encoder. As shown in Figure \ref{fig:attours}, unlike CoOp in Figure \ref{fig:head}(a), our method successfully inherits the original CLIP's focus on new classes and adjusts the attention to some extent, achieving the goal of preventing overfitting. Our method effectively achieves the overall transfer of the feature space to downstream tasks while inheriting the original CLIP's zero-shot knowledge.

\noindent\textbf{Hyperparameter sensitivity analysis:} 
\begin{itemize}
    \item \textbf{Mask Threshold.} Figure \ref{fig:ablations_onthers}(a) shows the impact of the Mask threshold on the harmonic mean accuracy. Overall, performance initially increases with a higher mask threshold but then decreases as the threshold continues to rise. Using a masking threshold of 30 provides the highest overall harmonic mean accuracy.
    \item \textbf{Constraint Parameters.} Figure \ref{fig:ablations_onthers}(b) and \ref{fig:ablations_onthers}(c) show the impact of ${\lambda}$ and ${{\gamma}}$ on the harmonic mean accuracy. Based on the overall model performance, we choose ${\lambda }$ = 1 and ${\gamma}$ = 3 as the hyperparameters for ${\mathcal{L}_{\rm{HL}}}$ and ${\mathcal{L}_{\rm{SP}}}$.
    \item \textbf{Prompt Depth.} In Figure \ref{fig:dpth} (left), we illustrate the effect of prompt depth in LPT and perform ablation experiments. Overall, performance improves as prompt depth increases. However, we observe that when randomly initialized prompts are inserted into deeper layers of a frozen model, where the feature space is already well-established, some performance degradation occurs. A similar trend has also been reported in MaPLe. Overall, LPT achieves optimal performance at a depth of 7.
    \item \textbf{Prompt Length.} Figure \ref{fig:dpth} (right) shows the effect of prompt length in LPT. As prompt length increases, the performance on base classes remains stable, while the accuracy on novel classes generally decreases. This indicates the emergence of overfitting, which negatively impacts generalization to novel classes.
\end{itemize}
\vspace{-2mm}
\section{Conclusion}
\label{sec:conclusion}
In this work, we introduce the LPT framework to tackle overfitting in base-to-novel tasks, improving generalization in VLM transfer. By filtering fine-grained foreground information, we reduce overfitting and enhance attention to underlying visual details. We also develop structural preservation~(SP) constraint loss and hierarchical logit~(HL) constraint to preserve CLIP's structural information while adapting to downstream tasks.  Our method shows significant improvements across 11 datasets and outperforms existing approaches.  Future work will focus on refining foreground filtering to minimize its impact on base class learning and developing more effective methods for capturing structural information to further address overfitting in VLM prompt tuning.

{
    \small
    \bibliographystyle{ieeetr} %
    \bibliography{main}
}


\end{document}